\documentclass{svproc}
\usepackage{float}
\usepackage{algorithm} 
\usepackage{graphicx}
\usepackage{amsmath,amssymb}
\usepackage{url}
\usepackage{algorithm}
\usepackage[noend]{algpseudocode}
\usepackage{xlop}
\usepackage{wrapfig,tabularx} 
\usepackage{booktabs}
\usepackage{graphicx}
\usepackage{xcolor}
\usepackage {longtable}
\usepackage[english]{babel}
\usepackage{array}
\usepackage{longtable}
\usepackage{booktabs}
\usepackage{multirow}
\usepackage{siunitx}
\usepackage{color, colortbl}

\usepackage{tikz}
\usetikzlibrary{positioning}
\usepackage{pgfplots}
\usepackage{microtype}

\def\checkmark{\tikz\fill[scale=0.4](0,.35) -- (.25,0) -- (1,.7) -- (.25,.15) -- cycle;}

\begin{document}
\mainmatter              
\title{Effect of Analysis Window and Feature Selection on Classification of Hand Movements Using EMG Signal}
\titlerunning{Effect of Analysis Window}  
%
%
%
\author{Asad Ullah\inst{1}\and
	Sarwan Ali \inst{1} \and
	Imdadullah Khan\inst{1} \and\
	Muhammad Asad Khan \inst{2} \and
	Safiullah Faizullah \inst{3},
}

\authorrunning{Asad  et al.} 

\institute{Department of Compute Science, Lahore University of Management Sciences (LUMS), Lahore, Pakistan \\
	\email{asaduet2662@gmail.com,\{16030030,imdad.khan\}@lums.edu.pk}
	\and
	Department of Telecommunication, Hazara University, \\ 
	Mansehra, Pakistan \\
	\email{asadkhan@hu.edu.pk}
	\and
	Department of Compute Science, Islamic University, \\
	Madinah, Saudi Arabia \\
	\email{safi@iu.edu.sa}
}

\maketitle              

\begin{abstract}
Electromyography (EMG) signals have been successfully employed for driving prosthetic limbs of a single or double degree of freedom. This principle works by using the amplitude of the EMG signals to decide between one or two simpler movements. This method underperforms as compare to the contemporary advances done at the mechanical, electronics, and robotics end, and it lacks intuition to perform everyday life tasks. Recently, research on myoelectric control based on pattern recognition (PR) shows promising results with the aid of machine learning classifiers. Using the approach termed as EMG-PR, EMG signals are divided into analysis windows, and features are extracted for each window. These features are then fed to the machine learning classifiers as input. By offering multiple class movements and intuitive control, this method has the potential to power Prosthesis an amputated subject to perform everyday life movements. In this paper, we investigate the effect of the analysis window and feature selection on classification accuracy of different hand and wrist movements using time-domain features. We show that effective data preprocessing and optimum feature selection helps to improve the classification accuracy of hand movements. We use publicly available hand and wrist gesture dataset of $40$ intact and $11$ trans-radial amputated subjects for experimentation. Results computed using different classification algorithms show that the proposed preprocessing and features selection outperforms the baseline and achieve up to $98\%$ classification accuracy for both intact and trans-radial subjects.
	
	%
\end{abstract}


\section{Introduction}
Biosignals are the electrical activity generated by an organ that represents a physical activity of human beings. These biosignals aided by the contemporary advancement in the fields of microelectronics and engineering are being used for various biomedical applications. More computationally efficient machines have seen an increasing trend in the interfaces based on these signals in brain-computer interaction (BCI) as part of the human-computer interface (HCI) \cite{merletti2004electromyography}. The most commonly used biosignals are Electroencephalography (EEG) and Electromyography (EMG) and Electrocardiography(ECG). These Biosignals, in general, and EMG signals, in particular, are used for the control of assistive devices for physically impaired people.  EMG are signals which are generated by the muscle contraction representing a neuromuscular activity \cite{reaz2006techniques,naik2014nonnegative}. These signals can be acquired either invasively or non-invasively. In the case of non-invasive EMG, called surface EMG (sEMG), are acquired by placing the electrodes on the skin of the subject (human).
In contrast, invasive signals are acquired intravenously using needle electrodes. Although invasive EMG is attenuated to certain problems that are associated with sEMG, such as muscle cross-talk, it has its shortcomings. These deficiencies include minor tissue damage and the reluctancy of people to the use of needles. The problem of muscle cross-talk can be accounted for by using better quality electrodes. 

Methods to operate prosthesis using myoelectric control has been around for more than five decades, such as proportionally control strategy. However, these devices offer a single or two degree of freedom (DoF) and offer restricted intuitive control \cite{lobo2014non}. More recently, Pattern recognition based EMG (EMG-PR) addresses the shortcomings of the proportionally controlled strategy, such as offering more hand and wrist movement and the dexterity involved with the conventional method \cite{samuel2019intelligent}.

\begin{figure}[H]
	\begin{center}
		\includegraphics[scale=0.65, page=2]{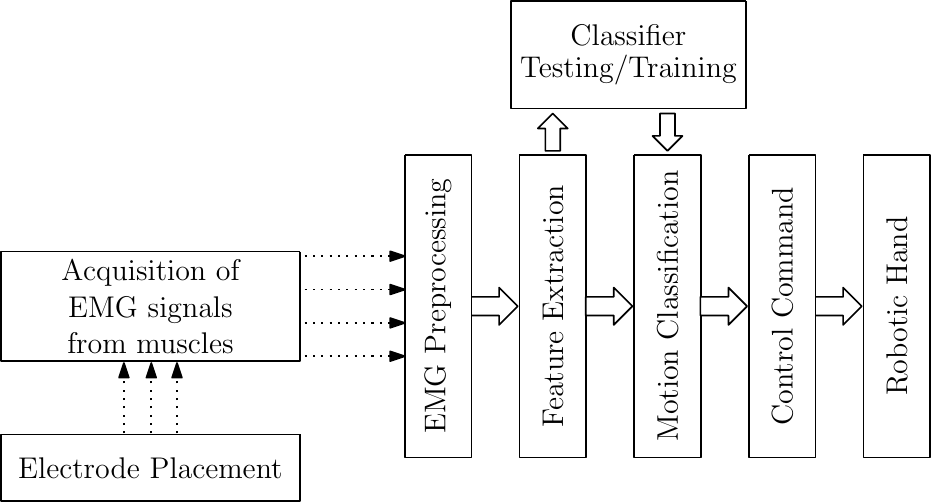}
		\caption{Flow chart of actuating robotic hand by sEMG signals}
		\label{fig_flow_chart}
	\end{center}
\end{figure}

EMG-PR based control strategy works on the principle that the EMG signals are distinct for a muscular activity mainly involving the upper limb. These signals are classified using machine learning techniques to decode the amputee's intended motion. The overall method is shown in Figure \ref{fig_flow_chart}. Despite the recent advances in signal classification for the activation of the prosthetic limb, this method is still not clinically viable. This is mainly due to the lack of optimal signal processing strategies, appropriate feature selection, and signal classification. The intricacies involved with the control system of EMG-PR often leads the patient to opt for the conventional prosthesis. About one-third of the amputees reject the self-powered prostheses \cite{atkins1996epidemiologic}.

The obstacle that is obfuscating the final goal of powering a prosthetic limb using sEMG signals is to find the optimum preprocessing and features selection that needs to be done to the signal to extract the features which can be feed to the classifier for accurate classification. Accurate prediction of hand movements involves fine data preprocessing, using appropriate analysis window length from which optimal features are extracted, and lastly, the choice of the classifier. Several works have explored different analysis window length and different kinds of preprocessing techniques, along with finding the optimal features to increase classification accuracy. Authors in \cite{zhai2016short} used principal component analysis for dimensionality reduction of the data. A window aggregation based approach is proposed in \cite{robinson2017pattern}, in which the authors aggregated five analysis windows from which features are extracted for classifying $17$ different movements of 11 subjects and achieved. A method based on different window lengths for classification of sEMG signals of 40 subjects is proposed in \cite{atzori2014electromyography}, which is evaluated on different datasets of EMG Signals for different hand and wrist movements. 

Designing a feature vector for different types of data is well explored in other domains as well. A network embedding approach is proposed in \cite{hassan2020estimating}, which converts the network/graph data into a fixed dimensional feature space while preserving its essential structural properties. A technique to compute similarity among graphs is proposed in \cite{shervashidze2009efficient}, which compute the similarity among graphs based on counting subgraphs (graphlets). A method to design feature vectors for attributes (properties) of nodes within a graph is proposed in \cite{ali2019predicting}. These feature vectors are given as input to different classification algorithms for prediction purposes. A method to measure the similarity between two data sequences is proposed in \cite{farhan2017efficient}, which can be used to classify the audio, images, and text sequences. A framework for sequence classification (music, and biological sequence classification etc.) is proposed in \cite{p2012generalized}, which uses general similarity metrics and distance-preserving embeddings with string kernels for classification.
 
In this paper, we investigate the effect of preprocessing of the raw data (signal) and variate the analysis window length to design a feature vector for the EMG signals. These feature vectors are then used for the classification of hand movements. We also investigate the impact of feature selection on efficiency and different combination of these features. We compare our propose technique with several baselines \cite{atzori2014electromyography,robinson2017pattern} and show that our proposed method outperforms them in terms of classification accuracy. For the baseline methods, we use the length of the analysis window and preprocessing techniques, as mentioned in the original studies. We apply four different classification algorithms to compute accuracies. Our main contributions are as follows: 

\begin{itemize} 
	\item We show that effective preprocessing helps to improve the classification accuracy of the underlying signal.
	\item We show that the optimum size of the analysis window is important to increase the overall performance of the approach.
	\item We show that selecting an optimum combination of features further helps to improve the performance.
\end{itemize}

The rest of the paper is organized as follows. In Section \ref{relatedwork}, we discuss the related work for EMG. 
Our proposed solution is given in Section \ref{solution}. Experimental settings, dataset description along with its preprocessing, and different configurations of feature is given in Section \ref{Experiments}. Results of our proposed method are given in Section \ref{results_and_comparisons}. We conclude the paper in Section \ref{Section_Conclusion}.

\section{Related Work}\label{relatedwork}
Previously, a dataset called Ninapro has been introduced by \cite{atzori2014characterization}, which was made public in the year $2014$. At the start, it had three databases, namely database 1 (DB1), database 2 (DB2), and database 3 (DB3). To date, it contains $8$ databases. This dataset is used in various works for the classification of hand movements. Authors in \cite{robinson2017pattern} used the database (DB2) for the first $11$ subjects to classify $17$ different hand and rest movements. They used a window size of $256$ms with an increment of $10$ms. They have used different configurations of seventeen time-domain features, with different classifiers (k nearest neighbor, random forest, and support vector machine). They reported that the random forest classifier achieves the highest accuracy, with an average classification accuracy of $90 \%$. They used a moving average for windows analysis in which the moving average of consecutive five windows. However, taking the average of multiple windows affected their classification accuracy because it obfuscates the important information in the signal. Also, as they only used data of $11$ subjects, which is much smaller. In a more recent work \cite{cene2019open}, data of $40$ subjects of the Ninapro dataset (DB2) was used. Four different time-domain features, with a window size of $200$ms along with an increment of $10$ms was used. The classifier ``Extreme leaning machine" was used for classification. Their average classification accuracy was $79 \%$. Although this work is comparatively better in terms of the number of subjects used, however, their selection of window size is not optimum. In a similar work \cite{zhai2017self}, authors used DB2 of the Ninapro dataset for all $40$ intact subjects and $50$ exercises, with a window size of $256ms$. They used a convolution neural network for classification and achieved an average classification accuracy of $78\%$. However, the runtime of the convolution neural network was the main problem with their approach.

In \cite{robinson2018effectiveness}, the authors used $11$ intact subjects of DB2 along with $9$ amputated subjects of the Ninapro dataset. They used the same analysis window configuration as that of the \cite{robinson2017pattern} and used time-frequency domain features by using the discrete wavelet transform of the data. They used the random forest algorithm for classification and achieved $90\%$ accuracy for intact and $75\%$ for amputated subjects. Since discrete wavelet features require the transformation of the signal, the approach proposed is computationally expensive. This is because the micro-controller embedded in the prosthetic limb has limited computational power, which makes this method obsolete in real-life usage. Authors in \cite{zhai2016short} used DB2 of the Ninapro data set. They used data of $40$ subjects for classification of sEMG signals using a spectrogram with an analysis window length of $200$ms and an increment of $100$ms. The spectrogram was computed using $256$ points Fast Fourier transform using a hamming window of $256$ points with an increment of $184$ points. To reduce the dimensions of the feature vector, they used principal component analysis (PCA) and achieved $75.74 \%$ average classification accuracy.  In \cite{atzori2014electromyography}, data of all $40$ subject is used  of DB2 for $50$ movements using the parameters used in \cite{gijsberts2014movement,englehart2003robust}. Their reported average classification accuracy is $75.27\%$. 

Authors in \cite{anam2013two} used two channels to classify combined finger movements, using the ``Extreme learning machine" for classification, with accuracy up to $98\%$. However, they have used only two surface electrodes, which make it less viable for clinical usage because two electrodes are not enough to record the motor activity.In \cite{tenore2008decoding}, they managed to achieve an accuracy of more than $90\%$ by using $32$ channels sEMG for classification of $10$ finger movements. In \cite{zhang2018combined}, the effect of various machine learning classifiers on the accuracy is analyzed. More recently, authors in \cite{jahan2015feature} used sEMG data of $48$ subjects for classification of movements. However, only four hand movements were considered in their experiments.

\section{Proposed Solution}\label{solution}
The sEMG signals are dynamic in nature (they vary quite fastly). Therefore, to encapture and analyze these signals, an overlapped window approach is used, which slides over the data with a predefined size and a preset increment $w$ (Figure \ref{Over_lapping_window}).  Characteristics of the window play an important role in feature extraction, which in turn affects the classification accuracy.

In our proposed approach, we used a window size of 256ms with an increment of $10$ms, which ensures that the threshold of $300$ms is not crossed. A window size greater than the threshold will induce a delay that a user might detect \cite{englehart2003robust}. The window size of $256$ ensures that a dense array of raw data is captured. Data samples corresponding to the size of $256$ms window length are stacked in a row vector for each of the electrodes, thus producing a segment of sEMG data. A total of $12$ sensors are used to record sEMG data. This results in a matrix of $ N\times(M*12) $ dimensions, where $N$ is the total number of windows, and $M$ is the number of sEMG voltage data samples that are in one window. Features were extracted from each of these windows $N$, resulting in a feature matrix $F$ of dimensions $ N\times P$, where $P$ is the number of features which in this work is $26$. This feature matrix is then given as input to the classification algorithms.

\subsection{Analysis Window}
\begin{figure}[h!]
	\begin{center}
		\includegraphics[scale=0.55,page=1]{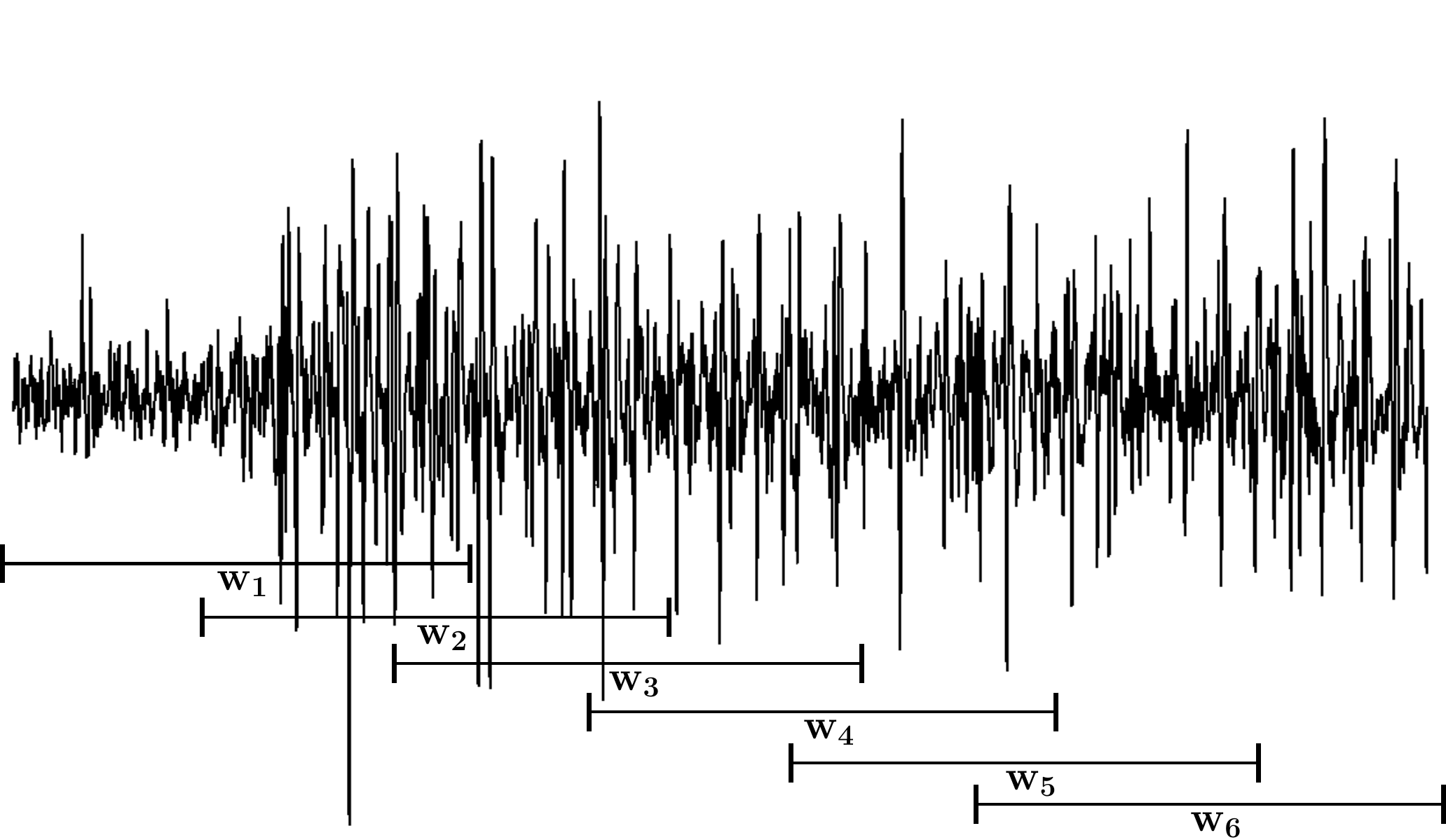}
		\caption{Sliding Windows}
		\label{Over_lapping_window}
	\end{center}
\end{figure}

Due to the random nature of the sEMG signal, its instantaneous value has almost no content to offer. For real-time analysis of the sEMG signal, same size of the analysis window is used as given in \cite{smith2010determining}. There are two kind of analysis windows, which are mostly used in pattern recognition based systems, $(i)$ adjacent window, and  $(ii)$ overlapping window. In $(i)$, a predefined length of windows is taken, and the next window starts at the end of preceding window. From these windows, features are extracted. However, it does not generate a dense array of signals, as in case of the adjacent window, the processing resources are underutilized.  In Overlapping window analysis, a predefined window length is taken, which slides over the signal with an increment size that is less than the original window size. This generates a dense array and makes full use of the available processing power \cite{Samuel2019IntelligentEP}. Figure \ref{Over_lapping_window} shows the overlapped window technique for a given EMG signal.

The size of the window has a direct relationship with accuracy. The greater window size is better to achieve higher classification accuracy. However, the greater window size is more expensive in terms of computational overhead.
The effect of window size is studied in \cite{scheme2011electromyogram}. Authors in \cite{scheme2011electromyogram} discover that the upper limit for the window size must be $300ms$. The optimal window size for the overlapped window technique has been investigated by various researchers and found that window size greater than $200ms$ is required to get higher accuracy \cite{smith2010determining}. 
In this paper, we used a window size of $256ms$ with an increment of $10ms$. This ensures that we have a dense array of windows from which we can extract features. 

\subsection{Feature Selection}
The problem in successfully implementing a myoelectric control prosthesis is that sEMG signals contain high dimensional data. To reduce the dimensions of data, features are extracted from the data, which can preserve the information of the signals. In literature, various time-domain features have been explored, which provides good classification accuracy. For instance, Integrated EMG (IEMG). Mean Absolute Value (MAV), Simple Square Integral (SSI), Variance (VAR) of EMG, Root Mean Square (RMS), Zero Crossing (ZC). Average Amplitude Change (AAC), and Willison Amplitude (WA). 

For our experiments, we extracted Zero Crossing (ZC), Mean Absolute Value (MAV), Slope Sign Changes (SSC), Mean Absolute Value Slope (MAV Slope), Waveform Length (WL), Root Mean Square (RMS), and $20$ frequency bins of histogram. A feature matrix (consist of $26$ features) is designed using these windows for each electrode. Thus a total of $312$ features were made for each movement. 

\subsection{Behavior of Features}
\begin{figure}[h!]
	\centering
	\includegraphics[scale=0.58, page=4]{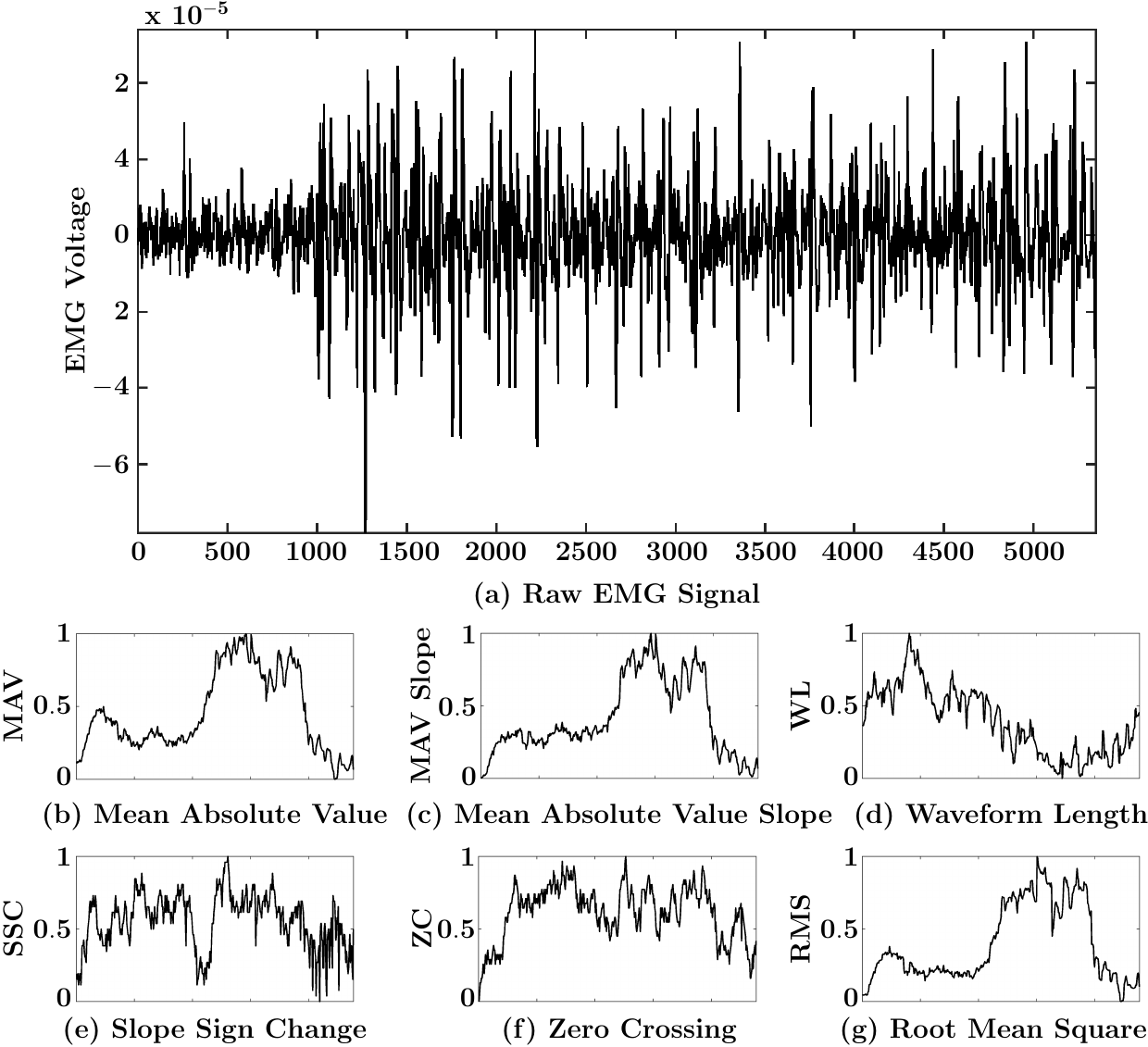} 
	\caption{Comparison of raw EMG signal for a hand movement with different features (normalized)}
	\label{plot_of_features}
\end{figure}

The choice of selecting a feature (or combination of features) depends on the behavior of the feature. For the purpose of choosing the optimal features, we analyze these features for different hand movements and compare them with raw EMG signals to understand the difference (see Figure \ref{plot_of_features}). Based upon the variation in the amplitude of the feature, we conclude that the lesser the overall gradient in the amplitude of the feature value, the better it would be for the describing EMG signal. Features also help in reducing the dimension of data. In Figure \ref{plot_of_features}, we plot different features (normalized) along with the raw EMG signal. We can observe in the figure that MAV, MAV Slope, and RMS are the features that do not have sudden changes in the amplitude and depicts uniformity in the course of a particular movement, while SSC and ZC have instantaneous changes in its amplitude. 

\subsection{Combinations of Features}
We use different configurations (combinations of features) in our experiments. Our goal is to find the best configuration so that dimensions of data can be decreased while increasing the classification accuracy. The configurations of features are given in Table \ref{tbl_featues_combination}. In configuration C1, we used all $26$ features, while different combinations of features are used in other configurations (from C2 to C7).

\begin{table}[h!]
	\centering
	\begin{tabular}{lccccccc}
		\hline
		Features & C1 &  C2 &  C3 &  C4 &  C5 &  C6 &  C7 \\
		\hline
		Mean Abs Value & \checkmark  &  &  &  &  &  & \checkmark \\
		Mean Abs Value Slope & \checkmark  &  &  &  &  &  & \checkmark \\
		Waveform Length & \checkmark  &  & \checkmark & \checkmark & \checkmark & \checkmark &  \checkmark \\
		Slope Sign Changes &  \checkmark &  &  &  &  & \checkmark & \checkmark \\
		Zero Crossing & \checkmark  &  &  &  & \checkmark &  & \checkmark \\
		Histogram & \checkmark  &  &  &  &  &  &  \\
		RMS & \checkmark  & \checkmark &  & \checkmark & \checkmark &  & \checkmark \\
		\hline
	\end{tabular}
	\caption{List of features and the different configurations ``C1 to C7" (combinations of features) used in our experiments}
	\label{tbl_featues_combination}
\end{table}

\subsection{Choice of Classifier}
Selection of classifier greatly affects the classification accuracy of EMG signals. In this paper, we consider four commonly used classifiers, namely k-Nearest Neighbors (kNN), Decision Tree (DT), Support Vector Machine (SVM), and Naive Bayes (NB).

\section{Experimental Evaluation}\label{Experiments}
In this section, we describe the dataset and baseline methods. The classification of hand movements is performed using R studio. All experiments are performed on the core $i5$ system with $4$GB memory.
\subsection{Dataset Description}
\begin{figure}[h!]
	\centering
	\includegraphics[scale=0.55,page=3]{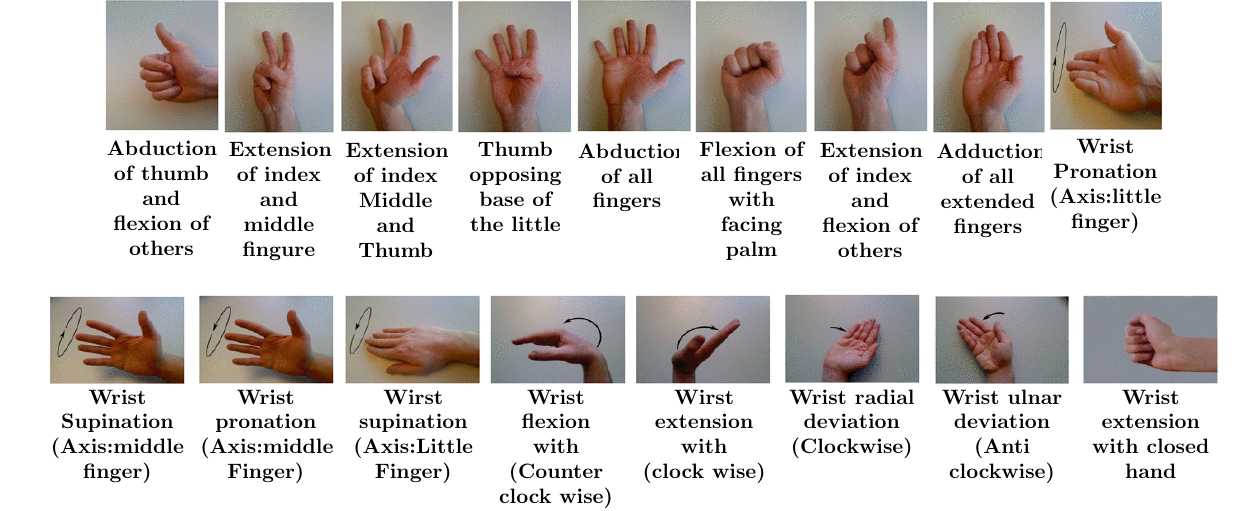}
	\caption{$17$ different hand and wrist movements}
	\label{fig_hand_movements}
\end{figure}

The data used in this paper is called Non-Invasive Adaptive Hand Prosthetics (Ninpro Project \footnote{\url{http://ninaweb.hevs.ch/}}) for sEMG signals for both intact and amputee subjects \cite{robinson2017pattern}. Currently, it contains $8$ datasets having data of $130$ subjects ($117$ intact subjects and $13$ trans-radial amputated subjects). Subjects perform series of wrist, hand, and finger movements (see Figure \ref{fig_hand_movements}). The data is generated in a controlled environment to avoid any noise. For experiments, we used DB2 and DB3 of Ninapro, which contains a total of $40$ intact and $11$ trans-radial subjects data. Intact subjects in DB2 performed a total of $50$ movements. These movements are further sub-categorized into E1, E2, and E3, each corresponding to different movement table. In this paper, we use E1, which contains data of $17$ movements shown in Figure \ref{fig_hand_movements}.

The acquisition setup for DB2 includes electrodes to record muscular activity. In the dataset, $12$ delsys double-differential sEmg electrodes with a base station were used to record sEMG signals. Out of the $12$ electrodes, eight were placed at a fixed radius around the forearm at the elevation of the radio-humeral joint. Furthermore, $2$ electrodes were placed on the biceps brachii and triceps brachii. Lastly, palpation was used to place the remaining two electrodes on the main activity spots to find the anterior and posterior of the forearm. After giving consent and physiological information such as age, height, and gender, each subject was asked to mimic a series of movements shown to them on a screen, which they performed $6$ times. The interval of each movement's repetition was $5$ seconds, followed by a rest period of $3$ seconds in which the subject was at rest state.

Ninapro Dataset contains DB3, which has data from eleven trans-radial amputated subjects. These amputated subjects performed the same $17$ hand movements, which were performed by intact subjects (see Figure \ref{fig_hand_movements}). The subjects performed these hand movements while imagining the posture of the wrist and finger for different tasks, and their sEMG signals were acquired.   

The data was sampled at a rate of $2kHz$. The raw sEMG signals are prone to noise (i.e., Power-Line noise), and therefore they were cleaned from $50Hz$. The data was relabeled to rectify the mismatching in the movement competition time.  Dataset statistics are given in Table \ref{tbl_Data_Set_Statistics}.

\begin{table} [h!]
	\centering
	\begin{tabular}{p{7cm} p{2.25cm} p{2cm}}
		\hline
		Statistics & Database $2$ & Database $3$ \\
		\hline
		Intact Subjects & 40 & 0 \\
		Trans-radial Amputated Subjects & 0 & 11  \\
		sEMG Electrodes & 12 Delsys & 12 Delsys \\
		Total Number of Movements (rest included)    & 50 & 50 \\
		Number of Movement Repetitions & 6 & 6  \\
		\hline
		E1 (Exercise B) ``Number of Movements" & 17 & 17 \\
		\hline
	\end{tabular}
	\caption{Ninapro Dataset Statistics}
	\label{tbl_Data_Set_Statistics}
\end{table}

\subsection{Baseline Methods}
For the first baseline method, we implemented the work of \cite{robinson2017pattern} with an overlapped window of size $256$ms and an increment of $10$ ms. Then average of $5$ consecutive windows are taken (see Equation \eqref{2017_paper_equation}). The disadvantage of taking the average of multiple windows is that it averaged out (distorted) the useful information in the signal, which in turn can affect the classification accuracy. We call this technique as ``Aggregation Window" (AG) technique.
\begin{equation}\label{2017_paper_equation}
WG_{t} = \frac{1}{n} \sum_{k=1}^{n} S_{nt-k}
\end{equation}
Where $n$ is the average of five windows, and $ S_{nt-k} $ is one window segment. 

The second approach, which we are using as a baseline, is proposed in \cite{atzori2014electromyography}. To implement this work, we used DB2 data of $40$ subjects. This data is segmented into the overlapped window of size $200$ms with an overlap of $10$ms. This approach does not aggregate the analysis window. Therefore, we call this technique as ``Without Aggregation" (WA) technique.

\subsection{Performance Measure}
To evaluate our proposed method, we compute accuracy for each subject (and each configuration) separately and compare them with the accuracies of the baseline methods. We also report average classification accuracy (average of every $10$ subjects) of our method and compare them with the average accuracies of the baselines.

\section{Results and Discussion}\label{results_and_comparisons}
In this section, we report the results of our proposed approach for different configurations and compare our results with different baseline methods proposed in the literature.
\subsection{Effect of features on classification accuracy}
We first evaluate the impact of different features on the classification of movement (see Figure \ref{plot_of_features} to observe the behavior of different features). 
Table \ref{feature_accuracy_comparison} shows the accuracy comparison of individual features for a randomly selected subject. We can observe in the table that MAV performs best in the case of kNN while WL performs best in the case of SVM classifier.
\begin{table}[h!]
	\centering
	\begin{tabular}{lp{1.5cm}p{0.75cm}}
		\toprule
		\multirow{2}{*}{Features} & 
		\multicolumn{2}{c}{Accuracy}  
		\\
		\cline{2-3} 
		& kNN & SVM \\
		\midrule
		
		Mean Absolute Value  & 93.1   & 72.7 \\
		Mean Absolute Value Slope & 90.4  & 68.1 \\
		Waveform Length  & 91.4  & 74.0 \\
		Slope Sign Change  & 67.9  & 56.7 \\
		Zero Crossing  & 74.4  & 59.3 \\
		Root Mean Square  & 90.8  & 66.4 \\
		\bottomrule
	\end{tabular}
	\caption{Accuracy (\%) comparison of different features on a randomly selected subject}
	\label{feature_accuracy_comparison}
\end{table}

\subsection{Intact Subjects Results}
\begin{figure}[h!]
	\centering
	\begin{tikzpicture}
	\begin{axis}[title={},
	compat=newest,
	xlabel style={text width=3.5cm, align=center},
	xlabel={{\small (a) Subjects 1 to 10}},
	ylabel={Percentage}, 
	ylabel shift={-7pt},
	height=0.43\columnwidth, width=0.55\columnwidth, 
	xtick={1,2,3,4,5,6,7,8,9},
	xticklabels={WA,AG,C1,C2,C3,C4,C5,C6,C7},
	legend columns=-1,
	legend entries={kNN,NB,DT,SVM},
	legend to name=CombinedLegendBar,
	]
	\addplot+[
	mark size=2pt,
	smooth,
	error bars/.cd,
	y fixed,
	y dir=both,
	y explicit
	] table [x={x}, y={knn}, col sep=comma] {10_avg.csv};
	\addplot+[
	mark size=2pt,
	mark=square*,
	dashed,
	error bars/.cd,
	y fixed,
	y dir=both,
	y explicit
	] table [x={x}, y={nb}, col sep=comma] {10_avg.csv};
	\addplot+[
	mark size=2pt,
	mark=diamond*,
	dotted,
	error bars/.cd,
	y fixed,
	y dir=both,
	y explicit
	] table [x={x}, y={dt}, col sep=comma] {10_avg.csv};
	\addplot+[
	mark size=2pt,
	dashdotted,
	error bars/.cd,
	y fixed,
	y dir=both,
	y explicit
	] table [x={x}, y={svm}, col sep=comma] {10_avg.csv};
	\end{axis}
	\end{tikzpicture}%
	\begin{tikzpicture}
	\begin{axis}[title={},
	compat=newest,
	xlabel style={text width=3.5cm, align=center},
	xlabel={{\small (b) Subjects 11 to 20}},
	ylabel shift={-3pt},
	yticklabels={},
	height=0.43\columnwidth, width=0.55\columnwidth, 
	xtick={1,2,3,4,5,6,7,8,9},
	xticklabels={WA,AG,C1,C2,C3,C4,C5,C6,C7},
	legend columns=-1,
	legend entries={kNN,NB,DT,SVM},
	legend to name=CombinedLegendBar,
	]
	\addplot+[
	mark size=2pt,
	smooth,
	error bars/.cd,
	y fixed,
	y dir=both,
	y explicit
	] table [x={x}, y={knn}, col sep=comma] {20_avg.csv};
	\addplot+[
	mark size=2pt,
	mark=square*,
	dashed,
	error bars/.cd,
	y fixed,
	y dir=both,
	y explicit
	] table [x={x}, y={nb}, col sep=comma] {20_avg.csv};
	\addplot+[
	mark size=2pt,
	mark=diamond*,
	dotted,
	error bars/.cd,
	y fixed,
	y dir=both,
	y explicit
	] table [x={x}, y={dt}, col sep=comma] {20_avg.csv};
	\addplot+[
	mark size=2pt,
	dashdotted,
	error bars/.cd,
	y fixed,
	y dir=both,
	y explicit
	] table [x={x}, y={svm}, col sep=comma] {20_avg.csv};
	\end{axis}
	\end{tikzpicture}%
	\\
	\begin{tikzpicture}
	\begin{axis}[title={},
	compat=newest,
	xlabel style={text width=3.5cm, align=center},
	xlabel={{\small (c) Subjects 21 to 30}},
	ylabel={Percentage}, 
	ylabel shift={-7pt},
	height=0.43\columnwidth, width=0.55\columnwidth, 
	xtick={1,2,3,4,5,6,7,8,9},
	xticklabels={WA,AG,C1,C2,C3,C4,C5,C6,C7},
	legend columns=-1,
	legend entries={kNN,NB,DT,SVM},
	legend to name=CombinedLegendBar,
	]
	\addplot+[
	mark size=2pt,
	smooth,
	error bars/.cd,
	y fixed,
	y dir=both,
	y explicit
	] table [x={x}, y={knn}, col sep=comma] {30_avg.csv};
	\addplot+[
	mark size=2pt,
	mark=square*,
	dashed,
	error bars/.cd,
	y fixed,
	y dir=both,
	y explicit
	] table [x={x}, y={nb}, col sep=comma] {30_avg.csv};
	\addplot+[
	mark size=2pt,
	mark=diamond*,
	dotted,
	error bars/.cd,
	y fixed,
	y dir=both,
	y explicit
	] table [x={x}, y={dt}, col sep=comma] {30_avg.csv};
	\addplot+[
	mark size=2pt,
	dashdotted,
	error bars/.cd,
	y fixed,
	y dir=both,
	y explicit
	] table [x={x}, y={svm}, col sep=comma] {30_avg.csv};
	\end{axis}
	\end{tikzpicture}%
	\begin{tikzpicture}
	\begin{axis}[title={},
	compat=newest,
	xlabel style={text width=3.5cm, align=center},
	xlabel={{\small (d) Subjects 31 to 40}},
	ylabel shift={-3pt},
	yticklabels={},
	height=0.43\columnwidth, width=0.55\columnwidth, 
	xtick={1,2,3,4,5,6,7,8,9},
	xticklabels={WA,AG,C1,C2,C3,C4,C5,C6,C7},
	legend columns=-1,
	legend entries={kNN,NB,DT,SVM},
	legend to name=CombinedLegendBar,
	]
	\addplot+[
	mark size=2pt,
	smooth,
	error bars/.cd,
	y fixed,
	y dir=both,
	y explicit
	] table [x={x}, y={knn}, col sep=comma] {40_avg.csv};
	\addplot+[
	mark size=2pt,
	mark=square*,
	dashed,
	error bars/.cd,
	y fixed,
	y dir=both,
	y explicit
	] table [x={x}, y={nb}, col sep=comma] {40_avg.csv};
	\addplot+[
	mark size=2pt,
	mark=diamond*,
	dotted,
	error bars/.cd,
	y fixed,
	y dir=both,
	y explicit
	] table [x={x}, y={dt}, col sep=comma] {40_avg.csv};
	\addplot+[
	mark size=2pt,
	dashdotted,
	error bars/.cd,
	y fixed,
	y dir=both,
	y explicit
	] table [x={x}, y={svm}, col sep=comma] {40_avg.csv};
	\end{axis}
	\end{tikzpicture}%
	\\
	\ref{CombinedLegendBar}
	\caption{Average classification accuracy of 10 subjects (in each plot) using different techniques and classifiers}
	\label{avg_comparisons_of_all_techniques}
	
\end{figure}
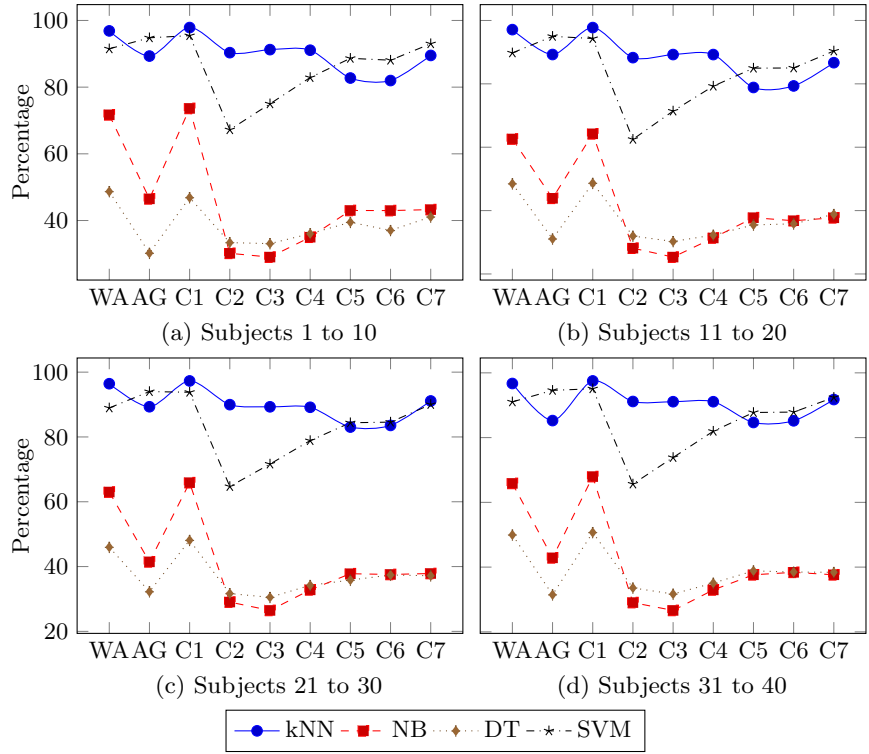

Average classification accuracy for all 40 subjects is shown in Figure \ref{avg_comparisons_of_all_techniques} for $10$ subjects each for our proposed methods and the baseline methods (Figure \ref{avg_comparisons_of_all_techniques} (a) shows average classification accuracy of subjects $1$ to subject $10$, (b) shows average over subjects $11$ to $20$,  (c) shows average over subjects $21$ to $30$ ) and (d) shows average over subjects $31$ to $40$ ) for $4$ classifiers namely kNN, NB, DT and SVM. The choice of classifiers has an important effect on the performance, as evident from the results. The SVM and kNN has an average classification accuracy higher than $85\%$ for all subjects and all approaches. The DT and NB are consistently underperforming for all methods and all subjects. Hence, their use in the classification of sEMG signals is obsolete.

In Figure \ref{avg_comparisons_of_all_techniques}, we can see that our proposed method (configuration C1) achieves better classification accuracy compared to the baselines. This behavior proves that the window size of $200$ is not optimum, as given by the WA approach. Aggregating the windows also showed to degrade the performance(as done by AG method). Thus it can be concluded that window size and averaging out windows effects the accuracy of the system. 

It is worth mentioning that the processors embedded in the microcontroller in the prosthesis have a limited amount of processing power. Therefore, to get a quick and real life like scenario at the cost of a few percent drops in the classification accuracy seems to be a choice that can be opted. In order to further reduce dimensions, we used other configurations, which contain a small set of features (see Table \ref{tbl_featues_combination}). It can be seen in Figure \ref{avg_comparisons_of_all_techniques} that although the configuration C1 outperforms all other configurations, C2, C3, and C4 with the number of features equal to $1$, $1$ and $2$ respectively have better accuracy compared to other configurations (with an average accuracy greater than $86\%$ in some cases). While configuration C5, C6, and C7 having features equal to $3$, $2$, and $6$, respectively, are underperforming in terms of accuracy compared to the others. Thus it is concluded that using more number of features does not necessarily produces better accuracy. This observation can be confirmed by analyzing Figure \ref{avg_comparisons_of_all_techniques}, which shows that configuration C2 having only a single feature has average classification accuracy better than C5, C6 and C7, which has more features compared to C2. Therefore, in a scenario where we can afford to drop the accuracy while increasing the computation speed, configuration C2, C3, and C4 can be a good choice to consider.

\begin{table}[h!]
	\centering
	\begin{tabular}{ccp{0.72cm}p{0.67cm}p{0.67cm}p{0.67cm}p{0.67cm}p{0.67cm}p{0.67cm}p{0.67cm}p{0.67cm}p{0.67cm}p{0.67cm}p{0.67cm}}
		
		\toprule
		\multirow{2}{*}{Classifier} &
		\multirow{2}{*}{Technique} &
		\multicolumn{12}{c}{Subjects} 
		\\
		\cline{3-14} 
		{} & {} & {1} & {2} & {3} & {4} & {5} & {6} & {7} & {8} & {9} & {10} & {11} & {12} \\
		\midrule
		\multirow{9}{*}{kNN} & WA  & 97.7 & 97.6 & 97.2 & 95.2 & 97.8 & 93.1 & 97.1 & 98.5 & 97.2 & 97.3 & 94.6 & 97.4 \\
		& AG & 90.2 & 87.9 & 89.6 & 85.6 & 89.3 & 88.8 & 87.8 & 91.0 & 92.7 & 90.4 & 88.4 & 89.9 \\
		& C1 & \textbf{98.3} & \textbf{98.1} & \textbf{97.7} & \textbf{96.2} & \textbf{98.3} & \textbf{97.2} & \textbf{97.5} & \textbf{98.6} & \textbf{98.2}  & \textbf{98.1}   & \textbf{95.9} & \textbf{97.9} \\
		& C2  & 92.7 & 88.9 & 94.1 & 83.8 & 96.3 & 89.0 & 82.1 & 89.6 & 94.6 & 92.3 & 87.3 & 89.9 \\
		& C3  & 92.2 & 89.4 & 93.5 & 86.4 & 94.9 & 89.5 & 85.8 & 90.7 & 94.2 & 95.9 & 85.6 & 91.5 \\
		& C4  & 92.1 & 89.1 & 93.0 & 86.4 & 95.0 & 89.0 & 85.4 & 90.6 & 94.3 & 95.9 & 85.6 & 91.2 \\
		& C5  & 82.2 & 84.3 & 83.6 & 84.4 & 87.1 & 87.5 & 65.8 & 76.1 & 87.6 & 88.8 & 77.6 & 80.1 \\
		& C6  & 81.9 & 85.8 & 86.0 & 85.2 & 87.3 & 87.3 & 64.3 & 70.5 & 85.4 & 86.1 & 82.4 & 77.2\\
		& C7  & 89.9 & 92.7 & 91.7 & 92.9 & 93.1 & 94.6 & 74.1 & 79.2 & 94.2 & 92.6 & 89.9 & 86.1\\
		\midrule
		\multirow{9}{*}{SVM} & WA  & 89.4 & 88.8 & 90.1 & 89.5 & 94.0 & 93.0 & 87.9 & 95.4 & 93.3 & 93.5 & 85.8 & 89.2\\
		& AG & \textbf{94.7}  & 93.9 & 93.6 & 92.9 & 95.6 & 94.6 &  \textbf{95.5}  & 96.0 & 96.0 & 95.1 & \textbf{93.9} & \textbf{95.2}\\
		& C1  & 93.9 &  \textbf{94.3} & \textbf{95.0} & \textbf{93.9} & \textbf{96.9} & \textbf{97.6}  & 93.1 &  \textbf{97.6}  & \textbf{96.8} & 96.3 & 92.2 & 93.3 \\
		& C2  & 73.8 & 62.6 & 74.3 & 56.0 & 79.0 & 63.6 & 48.2 & 64.5 & 77.6 & 72.7 & 54.9 & 67.2\\
		& C3  & 80.5 & 70.2 & 78.5 & 62.5 & 82.7 & 73.4 & 58.0 & 76.5 & 84.8 & 82.9 & 55.1 & 75.0\\
		& C4  & 86.0 & 80.4 & 85.0 & 72.8 & 88.0 & 79.2 & 70.2 & 87.4 & 90.4 & 89.3 & 66.4 & 83.7 \\
		& C5  & 90.1 & 87.2 & 88.9 & 82.6 & 92.2 & 89.0 & 76.0 & 93.0 & 93.3 & 94.1 & 75.0 & 87.2 \\
		& C6  & 90.4 & 87.7 & 89.1 & 79.6 & 92.8 & 86.5 & 76.1 & 92.3 & 93.2 & 93.3 & 76.1 & 88.3 \\
		& C7  & 93.3 & 93.6 & 93.5 & 87.7 & 95.6 & 93.0 & 83.8 & 96.6 &  {96.7}  & \textbf{96.4}  & 83.6 & 90.9\\
		
		\bottomrule
	\end{tabular}
	\caption{Accuracy (\%) comparison of existing approaches with our proposed setting using different classifiers}
	\label{tbl_accuracy_10}
\end{table}

In Table\ref{tbl_accuracy_10}, the individual accuracy of first 12 subjects is shown for all $9$ approaches. The bold values show the maximum accuracy. As it is evident from Figure \ref{avg_comparisons_of_all_techniques}, the use of Naive baise(NB) and decision tree(DT) classifiers results in accuracy, which is way below acceptance level so it is of no use to show their individual accuracy. So only kNN and SVM classifiers are shown. 

\subsection{Amputated subjects result }
In Table \ref{tbl_amputated_11}, we compare results for the eleven trans-radial amputated subjects (for $3$ different approaches, namely WA, AG, and configuration C1) using two different classifiers. Unlike intact subjects, SVM is the highest performing classifier in the case of amputated subjects.     
\begin{table}[h!]
	\centering
	\begin{tabular}{ccp{0.67cm}p{0.67cm}p{0.67cm}p{0.67cm}p{0.67cm}p{0.67cm}p{0.67cm}p{0.67cm}p{0.67cm}p{0.67cm}p{0.67cm}}
		\toprule
		\multirow{2}{*}{Classifier} &
		\multirow{2}{*}{Technique} &
		\multicolumn{11}{c}{Subjects} 
		\\
		\cline{3-13} 
		&  & {1} & {2} & {3} & {4} & {5} & {6} & {7} & {8} & {9} & {10} & {11}  \\
		\midrule
		\multirow{3}{*}{kNN} & WA  & 83.2 & 80.8 & 68.4 & 77.4 & 71.2 & 78.2 & 78.1 & \textbf{80.1} & 81.3 & 80.4 & 82.0  \\
		& AG & 79.5 & 77.0 & 57.7 & 56.8 & 63.9 & 66.1 & 67.1 & 77.7 & 84.3 & 77.5 & 80.6 \\
		& C1 & \textbf{86.5} & \textbf{84.5} & \textbf{73.9} & \textbf{86.3} & \textbf{77.8} & \textbf{83.1} & \textbf{83.2} & 77.7 & \textbf{85.3} & \textbf{84.8} & \textbf{85.8}  \\
		\midrule
		\multirow{3}{*}{SVM} & WA  & 89.5 & 93.0 & 78.2 & 78.0 & 97.2 & 87.0 & 58.8 & 91.8 & 95.9 & 91.4 & 95.4\\
		& AG & 89.5  & 88.7 & \textbf{90.3} & 88.2 & 90.8 & 82.7 &  \textbf{78.8}  & 88.7 & 93.9 & 89.9 & 93.4\\
		& C1  & \textbf{92.9} &  \textbf{95.7} & 84.6 & \textbf{92.9} & \textbf{98.3} & \textbf{91.0}  & 71.1 &  \textbf{98.5}  & \textbf{97.3} & \textbf{95.1} & \textbf{97.3} \\
		\bottomrule
	\end{tabular}
	\caption{Accuracy(\%) of eleven amputated subjects }
	\label{tbl_amputated_11}
\end{table}

\section{Conclusion and Future Work} \label{Section_Conclusion}
In this paper, we investigate the effect of the analysis window on the accuracy using the available sEMG dataset from the Ninapro database, which contains a variety of hand movements. Our results on different classifiers indicate that averaging out analysis window and the size of the analysis window affects the classification accuracy (because vital information gets averaged out). Furthermore, we also investigate the effect of different combinations of features on classification accuracy. We find out that kNN classifier has an average accuracy of more than $95\%$ with a particular configuration of features (C1).
In the future, we will use data that involves more hand and wrist movements. Furthermore, we will apply different dimensionality reduction based techniques (e.g. singular value decomposition) to reduce the dimensions of data while preserving valuable information. We will also try different methods to remove the noise from data so that the overall accuracy can be further improved.

\bibliographystyle{unsrt}
\bibliography{Effect_of_Analysis_Window_on_Classification_of_Hand_Movements_Using_EMG_Signal}

\end{document}